\documentclass{article}
\usepackage{bnaic}
\usepackage[pdftex]{graphicx}
\usepackage[cmex10]{amsmath}
\usepackage{amssymb}
\usepackage[caption=false,font=footnotesize]{subfig}
\usepackage{hyperref}

\newcommand{\pixels}{32}
\newcommand{\timesteps}{5000}

\title{\textbf{\huge Exploration and Exploitation in Visuomotor Prediction of Autonomous Agents}}

\author{Laurens Bliek \affila \affilb}

\date{\affila\ \textit{Delft University of Technology, Faculty of Electrical Engineering, Mathematics and Computer Science, Mekelweg 4, 2628 CD  Delft, The Netherlands}\\
    \affilb\ \textit{Almende B.V.,  Westerstraat 50,  3016 DJ Rotterdam, The Netherlands}}

\begin{document}

\ttl
\thispagestyle{empty}

\begin{abstract}
\noindent
This paper discusses various techniques to let an agent learn how to predict the effects of its own actions on its sensor data autonomously, and their usefulness to apply them to visual sensors. An Extreme Learning Machine is used for visuomotor prediction, while various autonomous control techniques that can aid the prediction process by balancing exploration and exploitation are discussed and tested in a simple system: a camera moving over a 2D greyscale image.
\end{abstract}

\section{Introduction}
% no \IEEEPARstart

If robots are ever supposed to work in real, complex and uncontrolled environments, they should be able to learn autonomously rather than be preprogrammed with all the information necessary for their specific task. One way to achieve this is by using prediction: making use of the available sensor values and motor commands to predict the sensor data of the future. If this can be learned succesfully, the robot can correctly predict the effect of its own motor commands on its sensor data. One consequence would be the ability to make a distinction between effects on a robot's sensors that are caused by itself or the environment, and those that are caused by others. For example, if a robot mounted with a camera would move to the left, it is expected that the camera pixels will move to the right, and after performing this command it can choose to perform this command again or choose to do something else. In the field of computer vision, the focus lies mainly on extracting information from the sensor data alone (like in object recognition), while the focus in this paper is to let an agent extract sensorimotor information autonomously. This can be done by making use of the known motor commands and the ability to control them. In theory, this should be possible without providing any of this information to the agent.

Of course, one issue is how to control these motor commands. In control theory, the goal is to choose those motor commands that will move a system towards a desired state by using a reference signal. Prediction techniques can be used to aid in this process. But for the problem described above, there is no external reference signal, since only autonomous agents are considered. The control problem is reversed: prediction (as defined in the next section) is the goal, while control can be used as an aid in the prediction process. How to control the motor commands in such a way that prediction is optimised, is still an open question, especially for visual sensors.

The problem can be divided into two parts: the prediction part and the control part. The control part also contains two subproblems, which can be seen as an exploration vs. exploitation trade-off: exploitation in this case is equivalent to moving towards predictable parts of the sensorimotor space so that prediction is performed correctly, while exploration is equivalent to moving towards new parts of the sensorimotor space so that the predictor can learn to predict there as well.

There has been some progress in this area in the past years, like the playful machines of Ralf Der and Georg Martius~\cite{der2012playful}, or the playground experiment of Kaplan and Oudeyer~\cite{oudeyer2005playground}. These techniques have been applied in both simulated and real robots, with impressive results. However, the sensors of these robots were usually low-dimensional, and the predictors were not always very complex. The contribution of this paper is to extend these techniques to agents with visual sensors and more complex predictors to cope with the dynamics of these systems. Visual sensors have the advantage that there is a direct effect of the motor commands on the sensor data, but a drawback is that visual data is in general high-dimensional.

In Section \ref{sec:pred} the prediction problem and the used prediction technique will be presented, while Section \ref{sec:contr} will focus on control techniques that are able to find a balance between exploration and exploitation in the context of prediction for autonomous robots. In Section \ref{sec:exp}, some of these techniques are applied in an experiment with a moving $\pixels \times \pixels$ pixel camera. Section \ref{sec:res} shows the results of these experiment, and Section \ref{sec:concl} concludes this paper.

% You must have at least 2 lines in the paragraph with the drop letter
% (should never be an issue)

\section{Prediction}
\label{sec:pred}

\subsection{Problem}

Suppose there is an agent with no knowledge about itself or its environment. It contains sensors and motors to interact with the environment, and some computational power to make calculations and store information. The goal of the agent is to `understand' its environment and the influence of its own motor commands on the sensor data. There is no clear definition of this understanding, but since prediction techniques will be used, the goal can be defined as follows: correctly predict the next sensor values $s_{t+1}$, given the current sensor values $s_t$ and motor commands $m_t$, for as much of the reachable sensorimotor space as possible. Here, $t$ is the timestep in discretised time, $S \subseteq \mathbb R^p$ is the $p$-dimensional sensor space, $M \subseteq \mathbb R^q$ is the $q$-dimensional motor space, and $s_t, s_{t+1} \in S$, $m_t \in M$. It is also assumed that $p>>q$.

Using this notation, prediction can be seen as a function $P: S\times M \rightarrow S$, and the mean-square prediction error $e_{t+1} = \frac{1}{p}||P(s_t,m_t) - s_{t+1}||^2$ should be minimised for multiple $(s_t,m_t)\in S\times M$. This section will discuss one technique to minimise this error, while being able to handle visual data. %Of course, dimensionality reduction techniques could be used to deal with this high-dimensionality. But ideally, the agent should discover the structures in its sensor data and the effects of its own motor commands on the sensor data by itself.

 The agent should not be told from the outside what is `good' or `bad': the sensor and motor data are all the information it has access to. Although no external signal is available, the prediction problem is still a supervised learning problem, since the desired output is available after one timestep. Not all supervised learning techniques are fit for this problem, because of the following properties:

\begin{itemize}
	\item Nothing is known about the properties of the data (like the noise and structure, or the relation between motor and sensor values).
	\item The mapping from input to output can be complex (nonlinear, non-deterministic, etc.).
	\item Both input and output are high-dimensional.
	\item Good generalisation is necessary for dealing with noise or with new situations.
	\item Learning needs to be done online.
\end{itemize}

One technique that can deal with most of these properties, is a relatively new neural network technique.

\subsection{Extreme Learning Machines}

\emph{Extreme Learning Machines}~\cite{huang2006extreme} (ELMs) are an efficient method for training a single hidden layer feedforward neural network. Instead of tuning the output and hidden layer by backpropagating the error, the hidden layer is initialised randomly and remains fixed during the learning process, while the output weights are the only adaptable parameters. Since the output neurons are chosen to be linear, this leads to a linear least-squares problem, while the nonlinear hidden neurons still allow the network to approximate nonlinear functions. The output weights can be computed by using a pseudo-inverse of the hidden layer output. 

The ELM algorithm can be summarised as follows:

\begin{enumerate}
	\item Given a training set $(x_t,y_t)$, with $x_t\in \mathbb R^n$ and $y_t \in \mathbb R^m$, $t = 1,\ldots, N$, initialise a single hidden layer feedforward neural network by choosing an activation function $g$, the  number of hidden neurons $\tilde N$, and weight matrices $W \in \mathbb R^{\tilde N \times n}$ and $\beta \in \mathbb R^{m \times \tilde N}$ that consist of the connections between input and hidden layer and between hidden and output layer respectively. A threshold $b \in \mathbb R^{\tilde N}$ can also be chosen for the hidden neurons. 
	\item Calculate the hidden layer matrix $H \in \mathbb R^{\tilde N \times N}$ consisting of the values of hidden neurons for each training sample by using $H^t = g(W x_t + b)$ for the columns of $H$.
	\item Keep the weights in $W$ fixed, but adapt the output weights $\beta$ by using the Moore-Penrose pseudo-inverse~\cite{rao1971generalized}: $\beta = Y H^\dagger$, with $Y$ the matrix consisting of desired output values $y_t$ in each column.
\end{enumerate}

The above algorithm gives the minimum norm least-squares solution to $\beta H = Y$, where $\beta H$ is the output of the neural network with fixed hidden layer parameters. Compared to traditional backpropagation algorithms, this  algorithm has several advantages:

\begin{itemize}
	\item Very large learning speed.
	\item Good generalisation performance because of the small output weights.
	\item Convergence to global minimum instead of possible local minimum entrapment.
	\item No need to adjust the learning rate.
\end{itemize}

Most of these advantages follow from the fact that there is a direct solution to the linear least-squares problem.

The current sensor values and motor commands of the agent can be used as an input of the ELM: $x_t = \left[\begin{array}{c} s_t\\ m_t\end{array}\right]$. If the next sensor value is chosen as the desired output of the ELM, i.e. $y_t = s_{t+1}$, the ELM can be used for visuomotor prediction. This gives:

\begin{equation}
	P(s_t,m_t) = \beta g(W \left[\begin{array}{c} s_t\\ m_t\end{array}\right] + b).
\end{equation}

\section{Control}
\label{sec:contr}

There are several possibilities for choosing a control that will aid the prediction process of the agent. The controller should make the agent explore its sensorimotor space while letting the predictor learn the relation between current sensor and motor values and the sensor values of the next timestep. As will be shown in this section, these are two contradictory goals, and several techniques for finding a balance between these goals will be discussed, as well as their applicability to agents with high-dimensional sensor data. The problem is closely related to the cognitive bootstrapping problem and the exploration vs. exploitation trade-off.

Note that no external reference signals are available. The goal of the controller is not to move the system towards a desired state, but to aid the prediction process. Therefore, only functions of internal signals can be used in the controller. To be more precise, the motor command $m_t$ at timestep $t$ is allowed to depend on the current sensor values and the past sensor and motor values only:
\begin{equation}
	m_t = C(s_t,m_{t-1},s_{t-1},m_{t-2},\ldots m_0,s_0).
\end{equation}

Since the prediction error  $e_{t} (s_{t-1},m_{t-1},s_t) = \frac{1}{p}||P(s_{t-1},m_{t-1}) - s_{t}||^2 $ also depends only on the current sensor values and the past sensor and motor values, it makes sense to use this error to choose the control. Several control techniques that can aid the prediction process are discussed below.

\subsection{Random movement}

To explore the sensorimotor space, the controller can choose random motor values:
\begin{equation}
 m_t \sim F_q,
\end{equation}
where $F_q$ is any $q$-dimensional probability distribution (continuous or discrete) that does not depend on the sensor or motor values. This technique is also called motor babbling. Given enough time, we can assume that the whole sensorimotor space will be explored just by performing random actions, if the probability for this is nonzero.

However, the curse of dimensionality prevents the practical use of this for high-dimensional problems, because the number of training samples required to get a good prediction for large parts of the sensorimotor space grows exponentially with the dimension of the space~\cite{bishop1995neural}. There is also no guarantee that the predictor can succesfully predict the result of a random movement. Besides that, random movements might cause the system to move towards states where no new information is provided to the predictor, reducing the efficiency and increasing the required number of timesteps to predict large parts of the sensorimotor space even more. The controller should aid the predictor by choosing actions efficiently. 
Nevertheless, interesting results have been obtained by combining this approach with chaotic search~\cite{tani1996model,tani1992proposal} or confidence~\cite{saegusa2009active}.

\subsection{Minimise prediction error}

By letting the controller choose those actions that minimise the prediction error, the controller can cooperate with the predictor, since now there are two subsystems that are adapted according to the same objective:
\begin{equation}
	m_t = \arg\min_M{ e_{t+1}(s_t,m_t,s_{t+1})}.
\end{equation}
Note that the prediction error $e_{t+1}$ is not yet known at timestep $t$, but it can be approximated by looking at past prediction errors.

This control technique should greatly increase the efficiency of the prediction process, and the prediction error is expected to decrease in a very short amount of time. However, the sensorimotor space will in general not be fully explored, and the predictor might only specialise in a small part of it (homeostasis), for example by predicting what will happen when the agent does not move. The controller should aid the predictor by providing exploratory behaviour so that the predictor can minimise the prediction error for a large part of the sensorimotor space.

\subsection{Maximise prediction error}\label{sec:maxPE}

To get exploratory behaviour, the controller could choose the actions in such a way that the prediction error is maximised:
\begin{equation}
	m_t = \arg\max_M{ e_{t+1}(s_t,m_t,s_{t+1})}.
\end{equation}
Since the predictor still minimises the prediction error, this approach causes the controller and predictor to be adversaries rather than to cooperate with each other. The controller makes sure the sensorimotor space is explored, while the predictor uses the new information in its learning process. This approach might work in theory, but several researchers~\cite{schmidhuber1991adaptive,schmidhuber1991curious,kaplan2004maximizing} pointed out that there are some drawbacks: since the controller is actually an adversary for the predictor, it might find ways to prevent the predictor from learning. For example, it might provide such instable motor behaviour that the sensor data becomes very noisy. Or it might move the system towards a part of the environment that is noisy. If it is possible to move the system towards a part of the sensorimotor space where no learning is possible, this approach can cause practical problems. The controller and predictor should not be adversaries, but complement each other, and there should be a balance between exploratory behaviour and predictability.

\subsection{Maximise learning progress}\label{sec:maxLP}

Kaplan and Oudeyer~\cite{kaplan2004maximizing} used reinforcement learning to drive an agent to situations where the learning progress was maximised. The learning progress is defined as the difference between the prediction errors of a few timesteps ago and the next prediction errors, for example:
\begin{eqnarray*}
	LP & = & em_{t} - em_{t+1},\\
	m_t & = & \arg\max_M{ LP },
\end{eqnarray*}
where $LP$ is the learning progress and $em_t$ is the mean of the errors between timesteps $t-w$ and $t$, for a certain sliding window $w$.

If the agent is learning, the prediction error is decreasing so there is a high learning progress. If the agent has learned its current situation, there is no progress, and if the agent cannot learn a specific situation there is also no progress. This solves the problem of staying in simple predictable or complex unlearnable situations. However, the new drawback is that if both simple and complex situations are available, the agent will alternate between them to let the error rise and drop everytime. The same authors have proposed solutions to this problem~\cite{oudeyer2004intelligent}.

\subsection{Other approaches}

The four approaches mentioned above are by far not the only ones. There are many more approaches to the problem of finding internal rewards that lead to a balance between explorative and predictive behaviour. An example is the playful machine of Ralf Der and Georg Martius~\cite{der2012playful}, where the concept of homeokinesis is introduced. Information-theoretic approaches also became popular over the past years~\cite{AyDer07}. An overview of both information-theoretic and predictive approaches concerning internal rewards is given in~\cite{oudeyer2007intrinsic}.

\section{Experiment}
\label{sec:exp}

\begin{figure}[!t]
	\centering
		\includegraphics[width=0.5\textwidth]{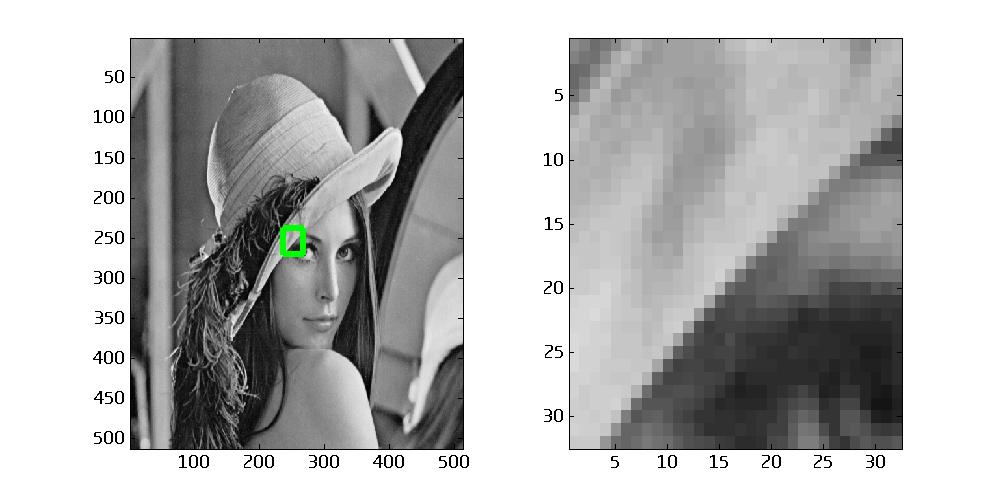}
	\caption{Experimental setup: a simulated $\pixels \times \pixels$ pixel camera can move over a $512\times 512$ greyscale image and has to learn the effects of its own movements on its sensor data using a neural network. The green rectangle on the left represents the part of the image that can be seen by the camera, the image on the right represents the output produced by the neural network after learning.}
	\label{fig:LenaCamera}
\end{figure}

To test whether the techniques discussed above can indeed be applied to agents with visual sensors, as a first experiment an active camera moving over a 2D picture has been simulated (see Figure \ref{fig:LenaCamera}). The picture contains $512\times 512$ greyscale pixels, the camera $\pixels \times \pixels$. The ELM technique from Section \ref{sec:pred} has been used for the predictor, in a neural network with $30$ hidden neurons. This number of neurons was found manually to give decent approximation capabilities while not requiring too much computation time, but it is not an optimal quantity. The predictor has to predict the next camera pixel values, given the current camera pixel values (with some added white noise) and the horizontal and vertical speed of the camera. For the control, the four techniques of Section \ref{sec:contr} were compared, with a controller that could choose the horizontal and vertical speed of the camera. The four techniques that were compared are: Random movement (RM), Minimise prediction error (MinPE), Maximise prediction error (MaxPE), and Maximise learning progress (MaxLP). The motor space was discretised into $5$ possible motor actions: move one pixel up, down, left or right, or stand still.

For the MinPE control, a simple method was used to minimise the error: at timestep $t$, look at the errors in the past up to a certain window $t-w$, and find the timestep $t^-$ where the error was minimal. At timestep $t$, the same motor command as the one used at timestep $t^-$ is chosen.
%\begin{equation}
%	m_t = \arg\min_{m_\tau\in\{m_{t-w},\ldots m_{t-1}\}} e_{\tau + 1}(s_\tau,m_\tau,s_{\tau +1}
%\end{equation}
 Besides this, at every timestep there is a probability of $P(\mathrm{random})$ that a random motor command will be chosen. A value of $P(\mathrm{random}) = 0.2$ has been chosen in this experiment. The MaxPE and MaxLP experiments use a similar approach. In the RM experiment, $P(\mathrm{random}) = 1$ and each motor command has the same probability of being chosen.
All experiments were run for $\timesteps$ timesteps.
%, which took about $20$ seconds for each of the considered techniques. To measure their performance, three different scores were defined that look at their ability to explore the sensorimotor space and their ability to predict (exploration vs. exploitation):
%
%\begin{itemize}
%	\item ExploitationScore: this is equal to the number of timesteps that the prediction error was less than a threshold of $0.1$, divided by the total number of timesteps ($400$).
%	\item ExplorationScore: this is equal to the number of unique points in sensorimotor space that appeared during the experiment, divided by the total number of timesteps.
%	\item TotalScore: this is equal to the number of unique points in sensorimotor space where the prediction error was less than $0.1$, divided by the total number of timesteps.
%\end{itemize}

\section{Results}
\label{sec:res}

%For the four chosen control techniques, the performance measures from the previous section were averaged over four experiments. Their values can be found in Table \ref{tab:results}, and an example of the camera movement can be found in Figure \ref{fig_routes}. The effects of choosing each control technique will be discussed in this section.

\begin{figure}[!t]
\subfloat[]{\includegraphics[width=0.25\textwidth]{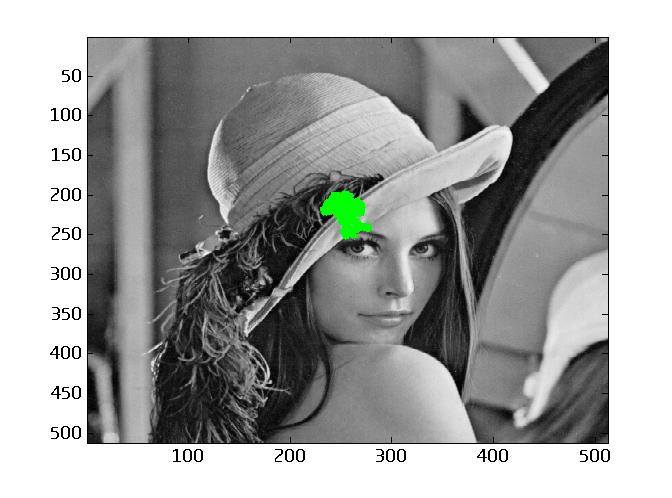}\label{fig_routeRM}}
\subfloat[]{\includegraphics[width=0.25\textwidth]{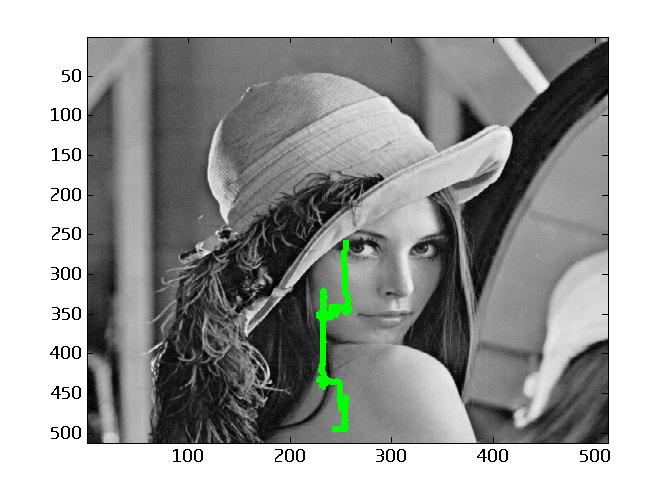}\label{fig_routeMinPE}}
\hfill
\subfloat[]{\includegraphics[width=0.25\textwidth]{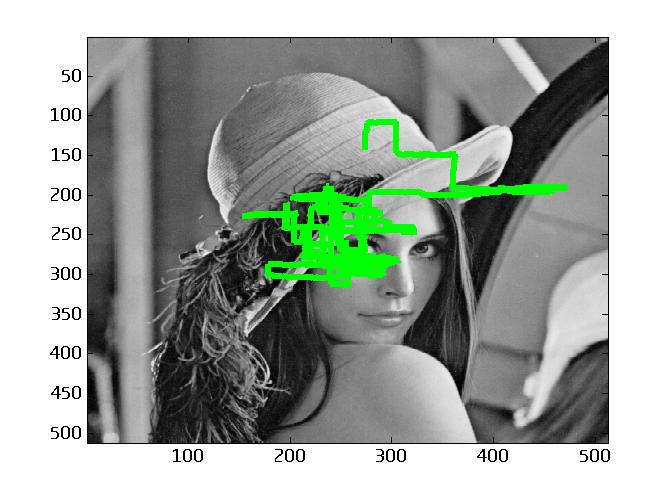}\label{fig_routeMaxPE}}
\subfloat[]{\includegraphics[width=0.25\textwidth]{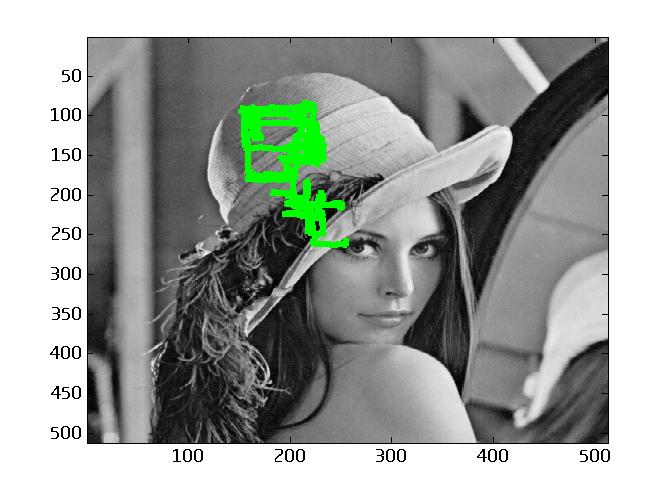}\label{fig_routeMaxLP}}
\caption{Movement of the center of the camera during $\timesteps$ timesteps of one simulation for four different control techniques: (a) Random Movement (RM), (b) Minimise Prediction Error (MinPE), (c) Maximise Prediction Error (MaxPE), (d) Maximise Learning Progress (MaxLP).}
\label{fig_routes}
\end{figure}

\begin{figure}[!t]
	\centering
		\includegraphics[width = 0.5\textwidth]{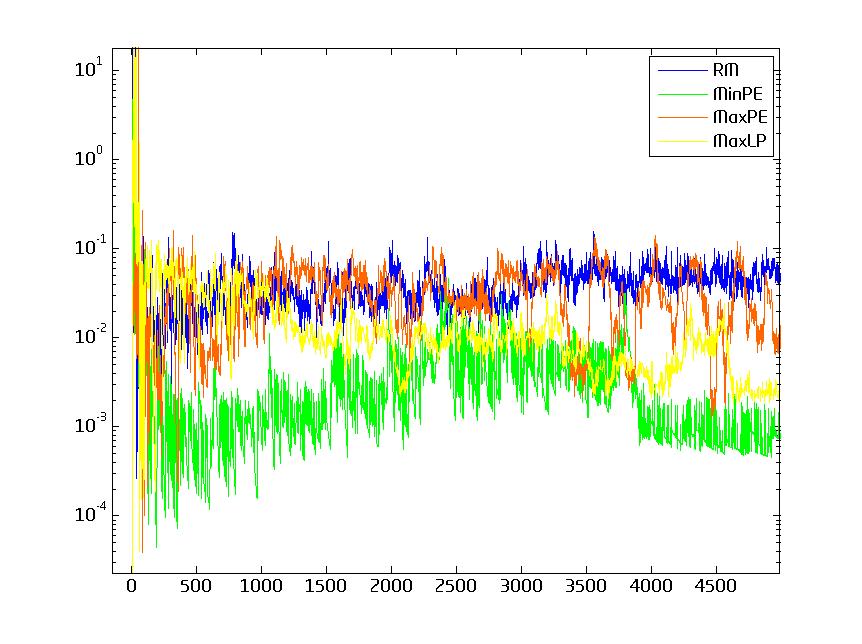}
	\caption{Prediction error during $\timesteps$ timesteps of one simulation of four different control techniques (RM, MinPE, MaxPE, MaxLP). Other simulations gave similar results. Note that the vertical axis has logarithmic scale. At the last timestep, RM has the highest error, followed by MaxPE, then MaxLP, and MinPE has the lowest error.}
	\label{fig_error}
\end{figure}

The explorative behaviour and the ability to predict were tested for the four different control techniques over several simulations. Most simulations of the same control technique gave similar results, and the outcome of one simulation for each technique is shown in Figure \ref{fig_routes}. The prediction errors can be found in Figure \ref{fig_error}. Since there is no clear metric for explorative behaviour, this has to be deduced from the camera movement. The camera movement for the four control techniques can be found in Figure \ref{fig_routes}. The effects of choosing each control technique will be discussed in this section.

In the RM approach, all motor commands were chosen with equal probability. Because of the symmetry in the available motor commands (move up, down, left, right, or stand still), this resulted in the camera moving around near its initial position. As such, only a small part of the sensor space was explored. The prediction error is higher than that of other approaches, probably due to the fast switching between commands.
 %Nevertheless this approach scores well because of the simplicity of the experiment: no complex motor commands were necessary to aid in the prediction process.

In the MinPE approach, only those motor commands were chosen that gave the smallest prediction error. In general, this caused the camera to %remain standing still, after choosing the initial motor commands for several timesteps.
repeat the same command for several timesteps, until another command would give a lower error. Since standing still usually gave the lowest error, this action was chosen a lot. This behaviour could be called homeostatic, and no explorative behaviour emerged. The prediction error was lowest in this approach, because mainly those actions were chosen that are easy to predict.
 %Because of this homeostatic behaviour, the ExploitationScore was really good for this technique, while the ExplorationScore was low.

The MaxPE approach did result in explorative behaviour. Large parts of the sensor space were explored. Standing still usually gave a low error, so this action was not chosen that often. The drawbacks mentioned about this approach in Section \ref{sec:maxPE} did not appear in this experiment because there were no unlearnable parts of the sensorimotor space: the prediction error was still low in most cases. Interestingly, the prediction error was lower than that of the Random Movement approach. This is probably due to the repeating of motor commands: the motor commands were chosen equal to the command that gave the highest error in the past, up to a certain timewindow. A different implementation of this approach might lead to a higher prediction error, especially if the environment contains unlearnable or very noisy parts.

%The MaxPE approach gave opposite results. One action was repeated until another action gave a higher prediction error, after which it chose to repeat this action. Standing still usually gave the lowest error, so this action was not chosen that often. This resulted in explorative behaviour and a good TotalScore. The drawbacks mentioned about this approach in Section \ref{sec:contr} did not appear in this experiment because there were no unlearnable parts of the sensorimotor space: the prediction error was below $0.1$ most of the time.

%\begin{table}[t]
%\caption{Performance of four different control techniques}
%\label{tab:results}
%	\centering
%		\begin{tabular}{|c|c|c|c|}
%		\hline
%		Control & ExploitationScore & ExplorationScore & TotalScore\\
%		\hline
%		\hline
%		RM & $0.94125$ & $0.68937$ & $0.65063$\\
%		\hline
%		MinPE & $\mathbf{0.96313}$ & $0.44813$ & $0.4225$\\
%		\hline
%		MaxPE & $0.92688$ & $0.71188$ & $0.65$\\
%		\hline
%		MaxLP & $0.93188 $& $\mathbf{0.75625} $& $\mathbf{0.7} $\\
%		\hline			
%		\end{tabular}
%		
%\end{table}

The MaxLP approach gave results similar to the MaxPE approach. The emerging behaviour was similar to the one in the MaxPE approach, i.e. exploratory. The prediction error was lower than that of the MaxPE approach, because the prediction error was no longer maximised, but it was not as low as in the MinPE approach. The result is a balance between exploration and exploitation. The experiment in this work was simple enough not to cause any of the drawbacks mentioned in Section \ref{sec:maxLP} about this approach. For more complex experiments, this approach might get stuck in alternating between highly predictable and unpredictable parts of the sensorimotor space, but here the ELM predictor was able to keep the prediction error low for most timesteps.
 %The emerging behaviour was similar to the one in the MaxPE approach, i.e. exploratory. The main difference was that when a certain motor command caused the prediction error to be large over multiple timesteps, the MaxLP approach caused the controller to choose a different command because of the lack of learning progress, while the MaxPE approach would keep generating this same motor command in this case. Therefore, the ExplorationScore and TotalScore of this approach were even better. The drawback mentioned in Section \ref{sec:contr} about alternating between unpredictable and easily predictable situations did not appear in this simple experiment, because the predictor was able to learn in all parts of the sensorimotor space.

Compared to other neural network applications, the speed of the simulations was very high. The simulations of this experiment with the ELM predictor only took about $6$ minutes for $\timesteps$ timesteps, while using the standard backpropagation algorithm would require at least several hours, and possibly would have a worse performance depending on the learning rate. For a smaller number of timesteps (around $400$), it was also possible to use the ELM technique in the same amount of time when the number of pixels was $300 \times 300$, which would never be possible with the backpropagation algorithm. Choosing a higher number of hidden neurons would probably result in a lower prediction error, but even with the small number of neurons used in this experiment, approximation was decent and the differences in the various control techniques were clear. Using more neurons, timesteps, and pixels usually resulted in memory problems.% In theory, the ELM should also be able to handle nonlinear sensorimotor dynamics, but this was not within the scope of this paper. %For more complex experiments, like dynamic environments, using ELMs might not suffice because it contains no temporal information.

\section{Conclusion}
\label{sec:concl}

In this paper, an Extreme Learning Machine was used for visuomotor prediction, and several control techniques that can aid in the prediction process were discussed. It was shown experimentally that at least some of these techniques can be applied to autonomous agents with visual sensors. Four control techniques were tested by applying them to a simulated $\pixels\times \pixels$ pixel camera that could autonomously control its horizontal and vertical movement. The results showed that these techniques can indeed let a high-dimensional system correctly predict its future sensor values, given its current sensor values and motor commands, while exploring the sensorimotor space, though only one of the tested techniques (MaxLP) provided a balance between exploration and exploitation. No information about the agent itself or the environment was necessary for the prediction process. Further research will show whether this can be extended to more complex experiments and real-world applications.

% conference papers do not normally have an appendix

% use section* for acknowledgement
\section*{Acknowledgment}

This research was performed at Almende B.V. in Rotterdam, the Netherlands. The author would like to thank Anne van Rossum for his supervision during this research, and Giovanni Pazienza for providing valuable feedback on this paper.

\bibliographystyle{plain}
\bibliography{mybiblio}

\end{document}